\pdfoutput=1

\documentclass[11pt]{article}

\usepackage[preprint]{acl}

\usepackage{times}
\usepackage{latexsym}

\usepackage[T1]{fontenc}

\usepackage[utf8]{inputenc}

\usepackage{microtype}

\usepackage{inconsolata}

\usepackage{graphicx}

\usepackage{microtype}
\usepackage{graphicx}
\usepackage{arydshln}
\usepackage{inconsolata}
\usepackage{todonotes}
\usepackage{tabularx}
\usepackage{booktabs} 
\usepackage{multirow}
\usepackage{amsmath}
\usepackage{amssymb}
\usepackage{graphicx}
\usepackage{subcaption}
\usepackage{fge}

\usepackage[framemethod=TikZ]{mdframed} 
\usepackage{listings}
\usepackage{longtable}
\usepackage{tcolorbox}
\usepackage{xspace}
\usepackage{wrapfig}
\usepackage{pgfplots}
\usepackage{color, colortbl}
\usepackage{array}
\usepackage{paralist}
\usepackage{geometry}
\usepackage{enumitem}
\usepackage[ruled,vlined,linesnumbered]{algorithm2e}
\usepackage{setspace}

\newcommand{\rparagraph}[1]{\vspace{1.2mm}\noindent\textbf{#1.}}

\definecolor{Gray}{gray}{0.92}
\definecolor{racing-green}{rgb}{0.0, 0.8, 0.6}
\definecolor{awesome-red}{rgb}{1.0, 0.13, 0.32}
\definecolor{LightCyan}{rgb}{0.88,1,1}
\definecolor{darkgreen}{RGB}{0,150,0}
\definecolor{Ground}{RGB}{255,184,55}
\definecolor{Dirt}{RGB}{191,169,115}
\definecolor{Pink}{RGB}{226,184,176}
\definecolor{Violet}{RGB}{163,148,170}
\definecolor{darkred}{RGB}{150,0,0} 
\definecolor{Red}{RGB}{171, 61, 56}
\definecolor{Green}{RGB}{62, 139, 117}
\definecolor{Blue}{RGB}{48, 110, 184}

\definecolor{CC}{RGB}{198, 226, 212} \definecolor{UU}{RGB}{198, 228, 253} 
\definecolor{CU}{RGB}{247, 202, 193} 
\definecolor{UC}{RGB}{242, 224, 253}

\newcommand{\ie}{\textit{i}.\textit{e}.,\ }
\newcommand{\eg}{\textit{e}.\textit{g}.,\ }

\definecolor{level4}{RGB}{110,136,203}
\definecolor{level3}{RGB}{173,190,226}
\definecolor{level2}{RGB}{205,208,243}
\definecolor{level1}{RGB}{236,236,252}

\newcommand{\cellcolorval}[1]{
   \ifdim#1pt>100pt \cellcolor{level4!90}#1\relax
   \else
   \ifdim#1pt>90pt \cellcolor{level4!90}#1\relax
   \else\ifdim#1pt>80pt \cellcolor{level3!70}#1\relax
   \else\ifdim#1pt>70pt \cellcolor{level3!70}#1\relax
   \else\ifdim#1pt>60pt \cellcolor{level2!40}#1\relax
   \else\ifdim#1pt>50pt \cellcolor{level2!40}#1\relax
   \else\ifdim#1pt>40pt \cellcolor{level2!10}#1\relax
   \else\ifdim#1pt>30pt \cellcolor{level2!10}#1\relax
   \else\ifdim#1pt>20pt \cellcolor{darkgreen!0}#1\relax
   \else \cellcolor{darkgreen!0}#1\relax
   \fi\fi\fi\fi\fi\fi\fi\fi
}

\newcolumntype{g}{>{\columncolor{Ground!7}}c}
\newcolumntype{d}{>{\columncolor{cyan!6}}c}
\newcolumntype{f}{>{\columncolor{lime!6}}c}
\newcolumntype{v}{>{\columncolor{purple!6}}c}
\newcolumntype{u}{>{\cellcolorval}c}

\title{\textit{All Roads Lead to Rome:} Graph-Based Confidence Estimation \\ for Large Language Model Reasoning
}

\author{
Caiqi Zhang$^{1}$\ \ 
Chang Shu$^{1}$\ \ 
Ehsan Shareghi$^{2}$\ \
Nigel Collier$^{1}$
\\
$^1$University of Cambridge \ \ \ 
$^2$Monash University \\
\texttt{\{cz391, cs2175, nhc30\}@cam.ac.uk} \ \ \ \texttt{ehsan.shareghi@monash.edu}
}

\begin{document}
\maketitle

\begin{abstract} 
Confidence estimation is essential for the reliable deployment of large language models (LLMs). Existing methods are primarily designed for factual QA tasks and often fail to generalize to reasoning tasks. To address this gap, we propose a set of training-free, graph-based confidence estimation methods tailored to reasoning tasks. Our approach models reasoning paths as directed graphs and estimates confidence by exploiting graph properties such as centrality, path convergence, and path weighting. Experiments with two LLMs on three reasoning datasets demonstrate improved confidence estimation and enhanced performance on two downstream tasks.
\end{abstract}

\section{Introduction}

Confidence estimation quantifies how certain a machine learning model is in its output. Higher confidence typically suggests a greater likelihood that the prediction is correct \citep{zhang2024atomic, logu, yang2025uncle}. This capability is critical in real-world applications, particularly in high-risk domains, where over-confidence can have serious consequences \citep{fadeeva-etal-2023-lm, zhang-etal-2024-luq, zhang2025reinforcement, zhang2025grace}. When confidence is low, alternative mechanisms (\eg model self-reflection or information retrieval) can be activated to enhance overall trustworthiness.

Most prior work on confidence and uncertainty estimation focuses on factual QA tasks \citep{kuhn2022semantic, xiong2023llms, fadeeva-etal-2023-lm, zhang-etal-2024-luq}. However, when applied to \textbf{reasoning tasks}, existing methods often fail due to two key differences between factual and reasoning QA: (1) model outputs in the latter are longer and include intermediate steps before the final answer; and (2) these intermediate steps are logically connected, unlike factual texts where facts can be verified independently \citep{zhang-etal-2024-luq, jiang2024graphbased}. Existing methods to estimate the confidence in reasoning tasks that focus \textbf{solely} on the final answer consistency \citep{wang-etal-2024-self-consistency, lyu2024calibrating} are thus insufficient. 

In this paper, we propose a series of novel, training-free, graph-based methods for confidence estimation in reasoning tasks. The graph structure can effectively capture the logical connections among different reasoning paths. Given a question and several independently sampled reasoning steps, we construct a directed graph to represent the reasoning process. Confidence estimation is then formulated using three graph-theoretic concepts: \textit{centrality}, \textit{path convergence}, and \textit{path weighting}. Our approach is model-agnostic and can be applied directly to any language model.

We evaluate our methods on two language models, Llama3.1-8B \citep{llama3modelcard} and Gemma2-9B \citep{team2024gemma}, using three reasoning-intensive benchmarks: MATH500 \citep{lightman2023lets}, MMLU-Pro \citep{wang2024mmlu}, and FOLIO \citep{han2022folio}. Our method consistently outperforms baseline approaches. To further demonstrate its effectiveness in downstream tasks, we apply it to selective self-reflection and LLM cascading, achieving improved overall performance with fewer reflection and cascading steps.

\section{Related Works}
\label{app:related}

\paragraph{Confidence Estimation in Reasoning.}
While most prior work on confidence estimation has focused on factual question answering, there is growing interest in applying similar techniques to reasoning tasks \citep{razghandi-etal-2025-cer, li2025language}. \citet{wang-etal-2024-self-consistency} and \citet{lyu2024calibrating} demonstrate that self-consistency can improve calibration in reasoning, though their approaches only consider final answers and ignore intermediate reasoning steps. The most similar  work to ours is CoT-UQ \citep{zhang2025cotuq}, which leverages self-prompt keyword extraction to enhance uncertainty quantification. While both methods address uncertainty in reasoning, our approach is from another angel, focusing on the structural graph properties of multiple reasoning paths rather than keyword extraction from a single path.

\paragraph{Graph-Based Confidence Estimation.}
Graph-based methods have been explored for confidence estimation in short-form QA \citep{li2024graphbased, lin2023generating}. In the context of long-form generation, \citet{jiang2024graphbased} construct semantic entailment graphs and use centrality measures to estimate confidence. Our approach differs fundamentally in both the graph construction methodology and our focus on complex reasoning tasks. \citet{yin-etal-2024-reasoning} propose token-level uncertainty as a guide for reasoning processes, while \citet{mo2023tree} introduce the Tree of Uncertain Thoughts, which employs Monte Carlo Dropout to estimate uncertainty at intermediate steps. Closest to our work, \citet{da2025understanding} also apply graph-based modeling to reasoning tasks; however, their graph construction strategy differs significantly from ours.

\paragraph{LLM-as-a-Judge and Process Reward Models.}
Another line of work focuses on verifying intermediate reasoning steps via \emph{process reward models} (PRMs) and \emph{LLM-as-a-judge} frameworks. These approaches train specialized reward models to assess the correctness of each step in a solution trajectory \citep{zhang2025lessons, lightman2023let, wang2023math, han2025verifiagent}. For example, \citet{lightman2023let} compare outcome vs.\ process supervision and introduce a PRM trained on 800K human-annotated step-wise feedback labels, significantly improving a model’s reliability on math problems. Similarly, \citet{zhang2025lessons} integrate an LLM-as-judge to label intermediate steps, finding that naive self-evaluation underperforms compared to using a large language model or human feedback. \citet{wang2023math} present \emph{Math-Shepherd}, a PRM trained with automatically generated step-level supervision signals for math, which is used to rerank solution steps and guide step-by-step reinforcement learning. More recently, \citet{han2025verifiagent} introduce a unified verifier, \emph{VerifiAgent}, that performs both meta-level consistency checks and tool-assisted verification, achieving broad improvements with fewer samples and lower cost than prior PRM-based methods. While effective in estimating the likelihood that each reasoning step is correct, these methods typically require heavy training and the use of specialized reward models, making them resource-intensive. Therefore, we do not include them in our comparisons, focusing instead on techniques that do not demand such dedicated reward model training.

\begin{figure*}[ht]
    \centering
\includegraphics[width=0.85\textwidth]{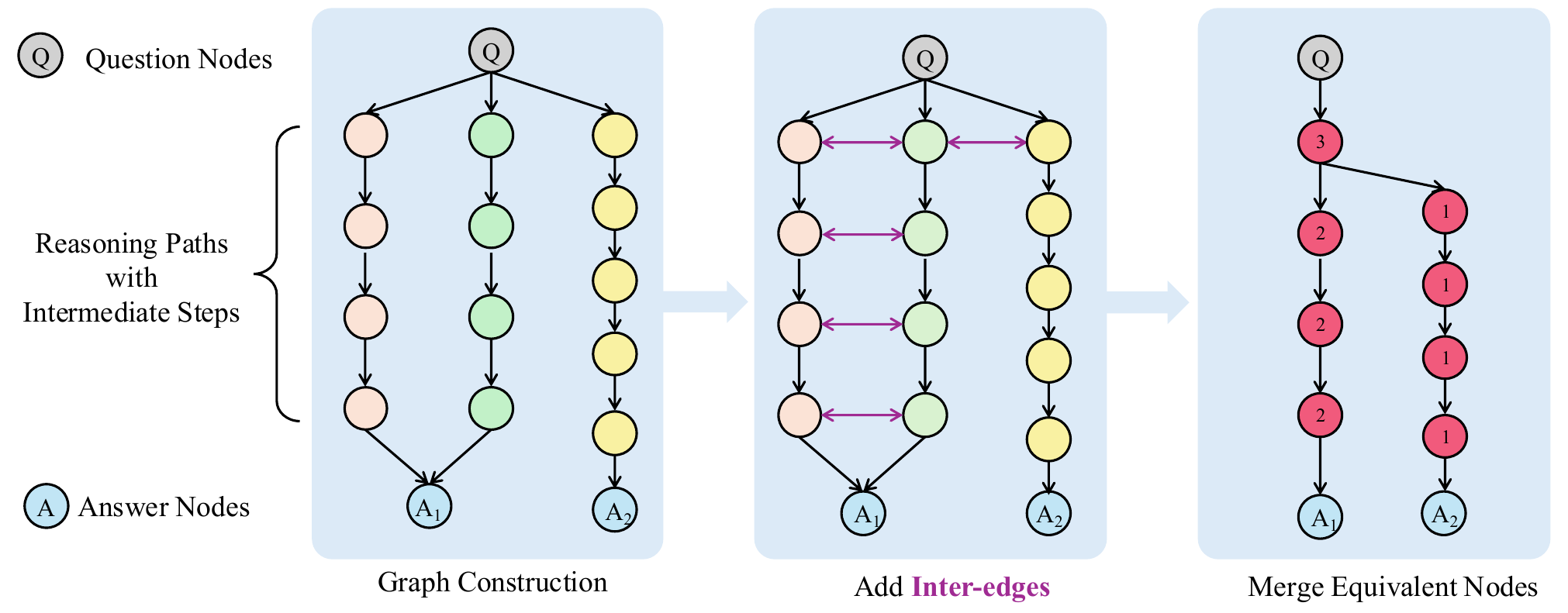}
\caption{Illustration of the graph construction. The left part shows the graph with only \textbf{\textit{intra-edges}}. The middle part includes \textbf{\textit{inter-edges}}. The right part shows the graph after merging equivalent nodes for the \textsc{PathWeight} method. Node weights are indicated inside the nodes.}
\vspace{-3mm}
    \label{fig:example}
\end{figure*}
\section{Methodology}

\paragraph{Motivation} Reasoning tasks can be solved through multiple pathways, each potentially leading to the same or different answers. This structure can naturally be modeled as a directed graph. The edges represent the logical flow of reasoning and nodes for each step. Each path from the question to an answer corresponds to a distinct reasoning route. Our central insight is that \textit{an answer supported by numerous, diverse paths is more likely to be correct}, reflecting the idea that ``all roads lead to Rome.'' Based on this intuition, our method generates multiple reasoning chains, merges them into a unified graph, and computes confidence from the resulting structure.

\subsection{Reasoning Chain Sampling} 
For a question $Q$, we generate $N$ reasoning chains by prompting the model, denoted as $\mathcal{C} = \{C_1, C_2, \dots, C_N \}$. Each reasoning chain $C_i$ can be further decomposed into a sequence of steps, \ie $C_i = \{s_{i1}, s_{i2}, \dots, s_{iT_i}\}$, where $T_i$ is the number of steps in the $i$-th chain. The final answers produced by the $N$ reasoning chains are denoted as $\mathcal{A} = \{A_1, A_2, \dots, A_N\}$. Identical answers across different chains are merged into a single node, resulting in a set of distinct answer nodes $\mathcal{A'} \subseteq \mathcal{A}$, with $|\mathcal{A'}|$ ranging from 1 to $N$.

\subsection{Graph Construction}
\label{sec:graph_construction}
We then construct a directed graph $\mathcal{G} = (\mathcal{V}, \mathcal{E})$ to represent the multi-path reasoning process. Each node corresponds to either a reasoning step or an answer, and edges capture the structure of reasoning. Starting from the question node $Q$, we iteratively add reasoning steps from each sampled chain. For each chain, we connect consecutive steps using \textit{intra-edges}, which are \emph{directed} edges representing the forward logical progression within a single reasoning path. In contrast, we also add \textit{inter-edges} between nodes from different chains that express equivalent meaning (identified via an auxiliary model; see Appendix~\ref{app:exp}). These \emph{bidirectional} inter-edges capture semantic equivalence across chains. The construction procedure is described in Algorithm~\ref{alg:graph-construction}. An simplified example with only three paths and a limited number of intermediate steps is shown in Figure \ref{fig:example}.

\begin{algorithm}[t]
\footnotesize
\caption{Graph Construction}
\label{alg:graph-construction}
\KwIn{Question $Q$ and reasoning chains $\{C_1, \dots, C_N\}$ where $C_i = \{s_{i1}, \dots, s_{iT_i}\}$}
\KwOut{Graph $\mathcal{G} = (\mathcal{V}, \mathcal{E})$}

Initialize $\mathcal{V} \gets \{Q\}$, $\mathcal{E} \gets \emptyset$\;

\For{$i \gets 1$ \KwTo $N$}{
  \For{$j \gets 1$ \KwTo $T_i$}{
    Add node $s_{ij}$ to $\mathcal{V}$\;
    \uIf{$j = 1$}{ \tcp{First node in the chain}
      Add edge $(Q \rightarrow s_{i1})$ to $\mathcal{E}$\;
    }
    \Else{
      Add intra-edge $(s_{i(j-1)} \rightarrow s_{ij})$ to $\mathcal{E}$\;
    }
  }
}

\ForEach{pair $(s_{ij}, s_{kl})$ from different paths}{
\If{$\text{Equivalent}(s_{ij}, s_{kl})$}{
    Add inter-edge $(s_{ij} \leftrightarrow s_{kl})$ to $\mathcal{E}$\;
  }
}
\For{$i \gets 1$ \KwTo $N$}{
  \If{$A_i \notin \mathcal{V}$}{
  \tcp{Only keep unique answers}
    Add answer node $A_i$ to $\mathcal{V}$\;
  } 
  Add edge $(s_{iT_i} \rightarrow A_i)$ to $\mathcal{E}$\;
}
\end{algorithm}

\subsection{Confidence Calculation}\label{sec:method}

Given a question $x$ and a model output $y$, the confidence is denoted as $\text{Conf}(x, y)$. 
With the graph $\mathcal{G}$, we formulate the confidence estimation problem in the following ways:

\rparagraph{Centrality-Based Confidence (\textsc{CenConf})}
Inspired by the self-consistency approach~\citep{wang-etal-2024-self-consistency, lyu2024calibrating}, which is essentially the \textit{in-degree centrality} of each answer node \citep{jiang2024graphbased}, we adopt \textit{Katz centrality} to capture more graph information.
Katz centrality evaluates a node's influence by considering both immediate neighbors and all other nodes that connect to it through these neighbors, with the influence of distant nodes attenuated by a factor~$\alpha$. Intuitively, an answer node that is reachable via numerous short, semantically meaningful paths is more likely to be correct. Further discussion on other centrality metrics is provided in Appendix~\ref{app:centrality}.

Formally, let $A$ denote the adjacency matrix of the graph $\mathcal{G}$. The Katz centrality score for node $v$ is defined as:

{\small
\[
\mathrm{Katz}(v) = \alpha \sum_{j} A_{vj} \, \mathrm{Katz}(j) + \beta
\]
}

where $\alpha$ is the attenuation factor (with $0 < \alpha < 1/\lambda_{\max}$, and $\lambda_{\max}$ being the largest eigenvalue of $A$), and $\beta$ is a constant representing the initial centrality assigned to each node. This recursive formula accounts for the influence of neighboring nodes, with the impact diminishing over longer paths.
After computing $\mathrm{Katz}(v)$ for each candidate answer node, we normalize the scores to obtain a confidence measure:

{\small
\[
\mathrm{Conf}(A_i) = \frac{\mathrm{Katz}(A_i)}{\sum_{A_j \in \mathcal{A'}} \mathrm{Katz}(A_j)}
\]
}

\rparagraph{Path Convergence Confidence (\textsc{PathConv})}  
In \textsc{PathConv}, we leverage the number of distinct reasoning paths from the question node \(Q\) to each candidate answer node \(A_i \in \mathcal{A'}\). The underlying intuition is that \textit{an answer reached by many distinct (or partially overlapping) paths is more likely to be correct}. 
Formally, we define: $
\mathrm{paths}(Q \to A_i) = \{\pi \mid \pi = (Q \to \dots \to A_i) \in \mathcal{G}\}.
$ The total number of paths that reach any answer is:

{\small
\[
P_{\text{all}} = \sum_{A_j \in \mathcal{A'}} \left| \mathrm{paths}(Q \to A_j) \right|
\]
}

The normalized confidence score is then computed as:

{\small
\[
\text{Conf}(A_i) = \frac{|\mathrm{paths}(Q \to A_i)|}{P_{\text{all}}}
\]
}





In cases where the graph \(\mathcal{G}\) is large and enumerating all paths is computationally intractable, we propose a path sampling strategy to approximate the path counts (Appendix \ref{app:pathconv}).

\rparagraph{Path Weighting Confidence (\textsc{PathWeight})}  
In \textsc{PathWeight}, we merge the nodes with equivalent semantic meaning.
Each merged node \(v\) is assigned a weight \(w(v)\) (default 1) representing the number of original steps combined. For a path \(\pi = (Q \to \dots \to A_i)\) traversing nodes \(\{v_1, v_2, \dots, v_k\}\), we define its score as:

{\small
\[
\mathrm{pathScore}(\pi) = \prod_{v \in \pi} w(v)
\]
}
The path weighting confidence is:
{\small
\[
\text{Conf}(A_i) = \frac{
  \sum_{\pi \in \mathrm{paths}(Q \to A_i)} \mathrm{pathScore}(\pi)
}{
  \sum_{A_j \in \mathcal{A'}} \sum_{\pi' \in \mathrm{paths}(Q \to A_j)} \mathrm{pathScore}(\pi')
}
\]
}
\textsc{PathWeight} emphasizes paths that incorporate commonly shared reasoning steps, thus boosting the confidence of answers supported by converging and repeated logic. Compared to simple path counting, the path weighting approach suppresses the influence of isolated or idiosyncratic steps, resulting in a more robust confidence estimation.

\begin{table*}[t]
\centering
\footnotesize
\setlength{\tabcolsep}{6pt}
\begin{tabular}{lccc ccc ccc}
\toprule
 & \multicolumn{3}{c}{\textbf{MATH500}} & \multicolumn{3}{c}{\textbf{MMLU-Pro}} & \multicolumn{3}{c}{\textbf{FOLIO}} \\
\cmidrule(lr){2-4} \cmidrule(lr){5-7} \cmidrule(lr){8-10}
\textbf{Method} & \textbf{AUROC$\uparrow$
} & \textbf{BS$\downarrow$} & \textbf{ECE$\downarrow$} & \textbf{AUROC$\uparrow$} & \textbf{BS$\downarrow$} & \textbf{ECE$\downarrow$} & \textbf{AUROC$\uparrow$} & \textbf{BS$\downarrow$} & \textbf{ECE$\downarrow$} \\
\midrule
\multicolumn{10}{c}{\cellcolor{gray!20} \textbf{Gemma-2-9B-It}} \\
\midrule
p(true) & 60.9 & 41.1 & 35.6 & 59.4 & 40.7 & 35.6 & 62.3 & 39.5 & 34.9 \\
Self-Verb & 61.6 & 36.8 & 33.5 & 60.7 & 37.7 & 32.8 & 64.4 & 35.7 & 31.2 \\
Self-Cons & 75.3 & 24.4 & 21.6 & 69.8 & 25.5 & 26.9 & 76.1 & 24.1 & 19.4 \\
Deg & 63.7 & 36.3 & 30.3 & 61.3 & 36.6 & 33.0 & 64.0 & 35.9 & 30.1 \\
Ecc & 64.9 & 34.6 & 31.0 & 62.9 & 35.7 & 31.7 & 68.1 & 35.1 & 27.6 \\
LUQ & 75.0 & 25.1 & 20.9 & 69.6 & 26.2 & 27.4 & 74.9 & 25.4 & 21.0 \\
\midrule
CenConf & \cellcolor{orange!20}76.7 & \cellcolor{orange!20}18.8 & \cellcolor{orange!20}17.0 & \cellcolor{orange!20}77.9 & \cellcolor{orange!100}18.3 & \cellcolor{orange!20}19.2 & \cellcolor{orange!20}78.3 & \cellcolor{orange!50}16.6 & \cellcolor{orange!20}16.7 \\
PathConv & \cellcolor{orange!50}80.9 & \cellcolor{orange!100}17.2 & \cellcolor{orange!50}16.6 & \cellcolor{orange!50}78.5 & \cellcolor{orange!50}18.7 & \cellcolor{orange!50}16.6 & \cellcolor{orange!50}81.2 & \cellcolor{orange!20}21.0 & \cellcolor{orange!100}11.2 \\
PathWeight & \cellcolor{orange!100}81.5 & \cellcolor{orange!50}17.8 & \cellcolor{orange!100}15.5 & \cellcolor{orange!100}80.8 & \cellcolor{orange!20}19.1 & \cellcolor{orange!100}13.1 & \cellcolor{orange!100}82.9 & \cellcolor{orange!100}15.4 & \cellcolor{orange!50}13.2 \\
\midrule
\multicolumn{10}{c}{\cellcolor{gray!20} \textbf{Llama-3-8B-Instruct}} \\
\midrule
p(true) & 64.0 & 36.3 & 31.2 & 63.3 & 37.4 & 32.5 & 64.8 & 35.1 & 30.6 \\
Self-Verb & 65.5 & 33.3 & 28.4 & 64.7 & 34.2 & 29.1 & 66.5 & 32.3 & 27.0 \\
Self-Cons & 79.3 & \cellcolor{orange!50}15.1 & 18.2 & 73.1 & 22.5 & 20.2 & 80.0 & \cellcolor{orange!20}20.5 & 17.4 \\
Deg & 68.2 & 32.2 & 27.1 & 66.3 & 33.1 & 28.3 & 70.2 & 30.3 & 25.2 \\
Ecc & 68.8 & 31.5 & 26.0 & 67.1 & 32.6 & 27.2 & 70.8 & 29.4 & 24.1 \\
LUQ & 79.5 & 17.5 & 17.5 & 73.5 & 23.2 & 19.5 & \cellcolor{orange!20}80.5 & 20.5 & 16.5 \\
\midrule
CenConf & \cellcolor{orange!20}83.0 & 16.1 & \cellcolor{orange!50}11.7 & \cellcolor{orange!20}81.0 & \cellcolor{orange!20}19.7 & \cellcolor{orange!20}16.3 & 80.5 & \cellcolor{orange!100}17.2 & \cellcolor{orange!20}13.2 \\
PathConv & \cellcolor{orange!50}83.5 & \cellcolor{orange!100}10.3 & \cellcolor{orange!20}15.8 & \cellcolor{orange!100}83.5 & \cellcolor{orange!50}15.5 & \cellcolor{orange!100}9.5 & \cellcolor{orange!50}84.0 & 21.8 & \cellcolor{orange!50}9.8 \\
PathWeight & \cellcolor{orange!100}84.0 & \cellcolor{orange!20}15.3 & \cellcolor{orange!100}10.4 & \cellcolor{orange!50}82.0 & \cellcolor{orange!100}15.1 & \cellcolor{orange!50}11.1 & \cellcolor{orange!100}85.5 & \cellcolor{orange!50}19.4 & \cellcolor{orange!100}9.5 \\
\bottomrule
\end{tabular}
\caption{Experiment Results on Gemma and Llama. All values are in percentages. The best three results are highlighted in \textcolor{orange}{orange}. The best results largely fall into our graph-based methods.}
\end{table*}

\section{Experiments}
\subsection{Setup}
\rparagraph{Models and Datasets}
We use the instruction-tuned Llama3.1-8B \citep{llama3modelcard} and Gemma2-9B \citep{team2024gemma}. For evaluation, we select MATH500 \citep{lightman2023lets}, MMLU-Pro STEM \citep{wang2024mmlu}, and FOLIO \citep{han2022folio}, which span arithmetic, STEM, and logical reasoning tasks.
We deliberately select tasks that yield a balanced mix of correct and incorrect predictions, as this scenario best showcases the value of confidence estimation.
\footnote{Datasets like GSM8K \citep{cobbe2021gsm8k}, where 7B-scale models already achieve >90\% accuracy, are less suitable for evaluating uncertainty estimation methods, since always predicting high confidence can still yield good calibration.}

\rparagraph{Baselines and Metrics}
Given the large number of confidence elicitation methods, we only select  representative approaches. We also exclude methods that cannot provide confidence to each response (\eg semantic uncertainty \citep{kuhn2022semantic}). For model self-reported confidence, we include: $p(\text{true})$ \citep{kadavath2022languagemodelsmostlyknow}, and self-verbalized confidence (Self-Verb) \citep{tian-etal-2023-just, xiong2023llms}. For consistency-based methods, we consider self-consistency (Self-Cons) \citep{wang-etal-2024-self-consistency, lyu2024calibrating}, Degree Matrix (Deg) \citep{lin2023generating}, and Eccentricity (Ecc) \citep{lin2023generating}. Additionally, we include LUQ \citep{zhang-etal-2024-luq}, a method specifically designed for long-form outputs. To ensure a fair comparison, we exclude PRMs \citep{wang2023math, zhang2025lessons}, as they require extensive training.
For evaluation, we use AUROC \citep{bradley1997use} as our main metric following \citep{kuhn2022semantic, lin2023generating}. For comparison, we also include commonly used metrics: Brier Score (BS) \citep{brier1950verification} and Expected Calibration Error (ECE) \citep{naeini2015obtaining}. 

\rparagraph{Experiment Settings}
We use NetworkX~\citep{SciPyProceedings_11} for graph construction and computation. The model is prompted to output in a clear structure, explicitly illustrating each step. Additional experiment details, evaluation strategies and prompts are provided in Appendix~\ref{app:exp} and \ref{app:prompt}.

\subsection{Main Results}

Across all three benchmarks and both foundation models, our proposed graph‑based methods consistently outperforms non‑graph baselines. 
For Gemma, \textsc{PathWeight} raised AUROC from 60.9\% to 81.5\% on MATH500 while cutting ECE from 35.6\% to 15.5\%; \textsc{PathConv} pushed the Brier Score down from 41.1\% to 17.2\% with a comparable AUROC of 80.9\%. Similar trends hold for Llama, where \textsc{PathWeight} reaches an AUROC of 84.0\% (vs.\ 64.0\%) and an ECE of 10.4\%, and \textsc{PathConv} attains the lowest Brier Score of 10.3\%. 
\textsc{CenConf} trails the two graph methods but still outperforms baseline methods. The results confirm that modeling logical connections and leveraging graph properties can significantly improve confidence estimation for reasoning tasks.

\section{Applications}
\label{sec:apps}

We demonstrate two downstream uses of the \textsc{Path‑Weight} confidence estimator: \emph{selective self‑reflection} and \emph{LLM cascading}.
Both interventions are triggered only for the $k \in \{5, 10, 15\}\%$ lowest‑confidence instances. 
For comparison, we also report a naive baseline that applies the same intervention to \emph{all} queries ($k = 100\%$).
Unless otherwise specified, the initial response is generated by Llama3.1‑8B-Instruct.

\rparagraph{Selective Self‑Reflection}
Self‑reflection prompts the \emph{same} model a second time to critique and revise its own answer.
For each low confidence example, we append a concise \textit{reflect‑then‑finalise} prompt (see Appendix~\ref{app:prompt}), asking the model to (i) identify flaws and (ii) provide a corrected response.

Table~\ref{tab:selfreflection} shows that reflecting on just the bottom 15\% of low-confidence queries yields accuracy improvements of $+3$ to $+5$ points across three benchmarks. 
Moreover, reflecting \emph{all} the time ($+100\%$) proves suboptimal—and in the case of MATH500, even \emph{reduces} accuracy.
This echoes prior findings that excessive self‑critique may cause correct answers to degrade, due to sycophancy \citep{sharma2023understanding} or over‑revision \citep{laban2023sure}.

\begin{table}[h]
\centering\footnotesize
\begin{tabular}{lccccc}
\toprule
\textbf{Dataset} & \textbf{Base} & \textbf{+5\%} & \textbf{+10\%} & \textbf{+15\%} & \textbf{+100\%} \\
\midrule
MATH500   & 49.8 & 52.1 & 53.4 & 54.3 & 53.0 \\
MMLU‑Pro  & 47.3 & 50.2 & 51.5 & 52.4 & 50.8 \\
FOLIO     & 63.5 & 66.0 & 66.8 & 67.2 & 64.8 \\
\bottomrule
\end{tabular}
\caption{Accuracy (\%) after Selective Self‑Reflection on the $k\%$ least‑confident samples.}
\label{tab:selfreflection}
\end{table}
\rparagraph{LLM Cascading}
Instead of reflecting, we can \emph{escalate} low‑confidence queries to a more capable (larger but slower) model.
In our setting, low-confidence cases are routed to Llama3‑70B-Instruct with the same prompt, while the rest are handled by the original Llama3‑8B-Instruct.
As shown in Table~\ref{tab:cascade}, cascading just the least-confident 15\% of queries yields accuracy improvements of around $+2$ to $+5$ points.

\begin{table}[h]
\centering\footnotesize
\begin{tabular}{lccccc}
\toprule
\textbf{Dataset} & \textbf{Base} & \textbf{+5\%} & \textbf{+10\%} & \textbf{+15\%} & \textbf{+100\%} \\
\midrule
MATH500   & 49.8 & 51.2 & 53.0 & 54.6 & 58.1 \\
MMLU‑Pro  & 47.3 & 49.1 & 51.7 & 52.9 & 56.3 \\
FOLIO     & 63.5 & 64.2 & 65.8 & 67.1 & 70.4 \\
\bottomrule
\end{tabular}
\caption{Accuracy (\%) when cascading the $k\%$ least‑confident queries to Llama3‑70B-Instruct.}
\label{tab:cascade}
\end{table}


\section{Conclusion}
We present a suite of graph-based, training-free methods for confidence estimation in reasoning tasks. By modeling reasoning paths and their logical connections as directed graphs, our approach captures deeper structural signals often overlooked by existing methods. Empirical results show consistent improvements over baselines across multiple benchmarks. Our methods also improve downstream applications such as self-reflection and LLM cascading, highlighting the practical benefits of our methods. We hope this work inspires future research into graph-based reasoning and uncertainty modeling in large language models.

\section*{Limitations and Future Work}

\textbf{Compute/latency overhead.} Our approach increases inference-time cost by sampling multiple reasoning chains and constructing a graph per instance. This cost is shared with consistency-style methods that similarly rely on sampling rather than internal logits. In practice, the overhead can be mitigated with early-exit heuristics (e.g., skipping graph construction when all samples agree) and by trading sample count for speed in settings with tight latency budgets.

\noindent \textbf{Black-box scope (no logits).} We intentionally design the method to be \emph{black-box} and model-agnostic, requiring only generated text. This broadens applicability to APIs and closed models where token-level logits are unavailable. A natural extension for \emph{white-box} access is to inject logit-derived signals into the graph: for example, (i) weight intra-edges in a chain by token- or step-level probabilities to reflect per-step confidence, and (ii) weight inter-edges by the verifier’s (or auxiliary judge’s) confidence in semantic equivalence. Such integrations could further sharpen both path aggregation and centrality scores, but would narrow the method’s deployment footprint to models exposing internals. We leave this to future work.

\noindent \textbf{Single-property estimators (no ensembling).} To isolate the contribution of distinct structural signals, we study three graph-based confidence estimators \emph{independently}. Prior work suggests that fusing complementary confidence cues can yield more robust estimators; ensembling these graph properties (e.g., via learned stacking, temperature-free weighted voting, or calibration-aware mixtures) is therefore a promising avenue we deliberately defer to future work.

\noindent \textbf{Sensitivity to graph construction.} Our estimates depend on faithful step decomposition and reliable cross-path equivalence detection. Noisy step segmentation or spurious equivalence links can perturb the topology (e.g., by creating shortcuts or cycles), which in turn can bias centrality and path counts. Robustness could be improved with stricter agreement thresholds, cycle-aware pruning, or by marginalizing over multiple equivalence graphs rather than committing to a single one.

\section*{Ethics Statement}
Our research adheres to strict ethical guidelines. We verified the licenses of all software and datasets used in this study to ensure full compliance with their terms. No privacy concerns have been identified. We have conducted a thorough assessment of the project and do not anticipate any further risks.

\section*{Acknowledgment}
We thank Chengzu Li and Ying Xu for their constructive feedback throughout this project. We are also grateful to the anonymous reviewers for their insightful comments and suggestions during the review process.

\bibliography{custom,anthology}

\clearpage
\newpage
\appendix
\section*{Appendix}

\section{More on Centrality-Based Methods}
\label{app:centrality}
During our experiment, we tested other centrality metrics such as closeness, pagerank, and Laplacian; however, they showed inferior performance compared to Katz centrality, which we utilize in CenConf. Katz centrality consistently yielded higher AUROC and significantly lower calibration errors, demonstrating superior effectiveness for this task.

\begin{table}[h]
\centering
\footnotesize
\setlength{\tabcolsep}{6pt}
\begin{tabular}{lccc}
\toprule
\textbf{Method} & \textbf{AUROC$\uparrow$} & \textbf{BS$\downarrow$} & \textbf{ECE$\downarrow$} \\
\midrule
Katz & 83.0 & 16.1 & 11.7 \\
Closeness & 69.2 & 22.5 & 15.6 \\
Pagerank & 65.3 & 24.6 & 18.2 \\
Laplacian & 64.3 & 27.3 & 19.2 \\
\bottomrule
\end{tabular}
\caption{Performance of Llama-3-8B-Instruct on MATH500.}
\end{table}

\section{More on \textsc{PathConv}}
\label{app:pathconv}

Algorithm~\ref{alg:path-sampling} outlines a randomized method that traverses the graph from \(Q\) up to a maximum path length \(L\), sampling \(M\) paths and accumulating weighted counts for each candidate answer \(A_i\).

\subsection{Path Sampling Strategy}  

\begin{algorithm}[h]
\footnotesize
\caption{Path Sampling for Approximating Weighted Path Counts}
\label{alg:path-sampling}
\KwIn{Graph \(\mathcal{G} = (\mathcal{V}, \mathcal{E})\), question node \(Q\), candidate answer \(A_i\), sample size \(M\), maximum path length \(L\), attenuation factor \(\gamma \in (0,1)\)}
\KwOut{Estimated weighted path count \(\widehat{P}(Q \to A_i)\)}
Initialize accumulator \(C \gets 0\)\;
\For{\(m \gets 1\) \KwTo \(M\)}{
    Set current node \(v \gets Q\)\;
    Set path weight \(w \gets 1\)\;
    \For{\(\ell \gets 1\) \KwTo \(L\)}{
        \textbf{if} \(v = A_i\) \textbf{then}\\
        \quad \(C \gets C + w\)\; 
        \quad \textbf{break} the current sample\;
        Obtain successor set \(\mathcal{N}(v)\) from \(\mathcal{G}\)\;
        \textbf{if} \(\mathcal{N}(v)\) is empty \textbf{then break}\;
        Sample a node \(v'\) uniformly from \(\mathcal{N}(v)\)\;
        Update path weight: \(w \gets w \times \gamma\)\;
        Set \(v \gets v'\)\;
    }
}
\Return \(\widehat{P}(Q \to A_i) \gets \frac{C}{M}\)\;
\end{algorithm}

The estimated weighted confidence for each answer \(A_i\) is then incorporated into the overall normalization:
\[
\text{Conf}(A_i) = \frac{\widehat{P}(Q \to A_i)}{\sum_{A_j \in \mathcal{A'}} \widehat{P}(Q \to A_j)}.
\]

\subsection{Avoiding Loops}
In our experiments, we observe that loops occasionally appear in the graphs (less than 5\% of cases). This is mainly due to misjudgments in equivalence checking. In such cases, we remove the loops by discarding certain inter-edges in the loops.

\section{Experiment Details}
\label{app:exp}
We use vLLM \citep{kwon2023efficient} for all model inference. 
To find equivalent steps, we prompt a Llama-3-8B-Instruct with temperature 0. 

For path generation, we use the temperature 1 with 3-shot in-context learning. We generate 10 additional samples for graph construction or consistency calculation. For fairness comparison, for p(ture) and Self-Verb, we directly ask the model whether each sampled answer is correct. For consistency-based methods, if the answer does not appear in the samples, it will be given a confidence 0.
For all methods, the output scores are normalized to the range \([0, 1]\) before computing ECE and Brier Score. 


\section{Hyperparameters }
\label{app:hyper-ablate}

\paragraph{Setup.}
Here we report the ablations conducted with \textbf{Llama-3-8B} on \textbf{MATH500}. Our default settings are $\alpha{=}0.1$, $N{=}10$ sampled traces, and $L{=}12$ maximum path length.

\subsubsection*{Attenuation Factor $\alpha$ (for CENCONF)}
Table~\ref{tab:alpha} shows that $\alpha{=}0.1$ yields the best AUROC with a favorable ECE, validating our default.

\begin{table}[h]
  \centering
  \small
  \caption{Ablation on attenuation factor $\alpha$ (CENCONF).}
  \label{tab:alpha}
  \begin{tabular}{lcc}
    \toprule
    \textbf{$\alpha$} & \textbf{AUROC (\%)} & \textbf{ECE (\%)} \\
    \midrule
    0.01            & 79.5 & 14.1 \\
    \textbf{0.1 (Default)} & \textbf{83.0} & \textbf{11.7} \\
    0.2             & 82.2 & 12.9 \\
    0.5             & 78.9 & 15.4 \\
    \bottomrule
  \end{tabular}
\end{table}

\subsubsection*{Number of Sampled Traces $N$ (for PATHWEIGHT)}
Increasing $N$ improves performance with diminishing returns (Table~\ref{tab:ntraces}). $N{=}10$ provides a strong cost--performance trade-off and is used as default.

\begin{table}[h]
  \centering
  \small
  \caption{Sensitivity to the number of sampled traces $N$ (PATHWEIGHT).}
  \label{tab:ntraces}
  \begin{tabular}{lcc}
    \toprule
    \textbf{$N$} & \textbf{AUROC (\%)} & \textbf{ECE (\%)} \\
    \midrule
    3                 & 79.1 & 16.2 \\
    5                 & 81.7 & 14.0 \\
    \textbf{10 (Default)} & \textbf{84.0} & \textbf{10.4} \\
    20                & 84.5 & 10.1 \\
    \bottomrule
  \end{tabular}
\end{table}

\subsubsection*{Maximum Path Length $L$ (for PATHCONV)}
We observed average reasoning chains of roughly 5--6 steps. Table~\ref{tab:maxlength} shows that $L{=}12$ avoids truncation errors seen at $L{=}5$; larger caps ($L{\ge}15$) yield no meaningful AUROC gains and only modest ECE improvements, so $L{=}12$ remains our default for efficiency.

\begin{table}[h]
  \centering
  \small
  \caption{Impact of maximum path length $L$ (PATHCONV).}
  \label{tab:maxlength}
  \begin{tabular}{lcc}
    \toprule
    \textbf{$L$} & \textbf{AUROC (\%)} & \textbf{ECE (\%)} \\
    \midrule
    5                 & 81.2 & 21.5 \\
    \textbf{12 (Default)} & \textbf{83.5} & 15.8 \\
    15                & 83.6 & 13.4 \\
    20                & 83.6 & 12.6 \\
    \bottomrule
  \end{tabular}
\end{table}

\clearpage
\newpage
\onecolumn
\section{Prompt}
\label{app:prompt}
\begin{table*}[h]
\centering
\vspace{-3mm}
\begin{tcolorbox}[colback=blue!5!white, colframe=blue!50!black, title = {\textsc{Prompt for Path Generation}}]
\ttfamily

Answer the following question. Break down your reasoning process into the smallest possible steps. Each step should represent a single, minimal reasoning action, and each step must logically follow the previous one. Use the following format for each step:

Step N:
Thought: [Provide a detailed explanation of your reasoning for this step.]

Present your entire reasoning process in one cohesive response.

After completing all the steps, conclude with:

Final Answer: \texttt{\textbackslash boxed\{[Your final numerical answer here without the unit or any additional text]\}}

Ensure that your response strictly follows this format to maintain clarity and consistency.

Question: \{question\}

\end{tcolorbox}
\vspace{-3mm}
\caption{Prompt for Path Generation.}
\label{tab:gen_binary}
\end{table*}

\begin{table*}[h]
\centering
\vspace{-3mm}
\begin{tcolorbox}[colback=blue!5!white, colframe=blue!50!black, title = {\textsc{Selective Self‑Reflection Prompt}}]
\ttfamily

Question: \{question\}

Previous Answer: \{model's earlier response\}

Please review your previous answer. Identify any errors or flaws, and revise your answer if necessary. For you final answer, output it in the following format:

Final Answer: \texttt{\textbackslash boxed\{[Your corrected final answer here]\}}

\end{tcolorbox}
\vspace{-3mm}
\caption{Prompt for Self‑Reflection.}
\label{tab:self_reflection}
\end{table*}

\begin{table*}[ht!]
\centering
\vspace{-3mm}

\begin{tcolorbox}[colback=blue!5!white, colframe=blue!50!black, title = {\textsc{Equivalence Checking Prompt}}]
\ttfamily

You are tasked with identifying the equivalent reasoning step in Path B for a specific reasoning step in Path A.

\textbf{Context:}  
Path A and Path B are sequences of reasoning steps.

\textbf{Inputs:}  

Target Step: \{A Step in Path A\}  

Steps in Path B:  \{All Steps in Path B\} 

\textbf{Your Task:}  
Identify the single step number in Path B that is equivalent to the given step in Path A. The equivalent step must:
\begin{itemize}
  \item Contain the same reasoning as the given step in Path A.
  \item Not contain additional or conflicting information.
\end{itemize}

If no such step exists, respond with \texttt{"none"}.

\textbf{Output Format:}  
Provide only the step number (e.g., \texttt{"5"}) or \texttt{"none"}.

\end{tcolorbox}
\vspace{-3mm}
\caption{Prompt for Equivalence Checking.}
\label{tab:path_alignment}
\end{table*}

\end{document}